\documentclass[10pt]{article} 

\usepackage{enumitem}
\usepackage[utf8]{inputenc} 
\usepackage{graphicx} 
\usepackage{amsmath} 
\usepackage{hyperref} 
\usepackage{algorithm2e}
\usepackage{multicol}
\usepackage{tabularx}
\usepackage{url}
\usepackage{caption}

\usepackage{cite} 

\usepackage[a4paper, total={7in, 10.75in}]{geometry}

\title{MRI Brain Tumor Detection with Computer Vision} 
    
\author{Jack Krolik, Jake Lynn, John Henry Rudden, Dmytro Vremenko} 
\date{\href{https://github.com/jack-krolik/brain-cancer-classifier-segmentation}{https://github.com/jack-krolik/brain-cancer-classifier-segmentation}} 

\begin{document}

\maketitle 
\begin{abstract}
This study explores the application of deep learning techniques in the automated detection and segmentation of brain tumors from MRI scans. We employ several machine learning models, including basic logistic regression, Convolutional Neural Networks (CNNs), and Residual Networks (ResNet) to classify brain tumors effectively. Additionally, we investigate the use of U-Net for semantic segmentation and EfficientDet for anchor-based object detection to enhance the localization and identification of tumors. Our results demonstrate promising improvements in the accuracy and efficiency of brain tumor diagnostics, underscoring the potential of deep learning in medical imaging and its significance in improving clinical outcomes.
\end{abstract}

\setlength{\parskip}{0pt plus 1pt}

\begin{multicols}{2}

\section*{Introduction}
Advancements in medical imaging and computer vision have paved the way for significant improvements in diagnostic methodologies. In particular, the detection and segmentation of brain tumors from MRI scans have seen transformative developments through the application of deep learning technologies. This paper examines various machine learning models tailored to enhance the accuracy and efficiency of brain tumor analysis, thereby aiding faster and more reliable medical diagnoses.

\section*{Problem Statement}
Brain tumors vary widely in size, shape, and location, making their detection and segmentation challenging yet critical for effective treatment planning. Traditional methods often require extensive manual review by radiologists, which is time-consuming and prone to human error. The challenge lies in developing robust automated systems that can accurately identify and categorize brain tumors from MRI scans, thus supporting radiologists in making timely and accurate assessments. This study aims to address these challenges by leveraging and optimizing state-of-the-art deep learning models for brain tumor detection and segmentation.

\section*{Data}
\subsection*{Brain Tumor MRI Dataset}

The Brain Tumor MRI Dataset comprises 7,023 MRI images sourced from figshare, SARTAJ, and Br35H datasets, segmented into four classes: glioma, meningioma, no tumor, and pituitary. For the binary classification task, we simply combined all tumor labels under the unified “Tumor” class \cite{d1}. 

\columnbreak

\begin{tabularx}{\linewidth}{|c|X|X|X|X|} 
\hline 
Split & No-Tumor & Meningi- oma & Glioma & Pituitary \\ \hline 
Train & 1,595 (27.92\%) & 1,339 (23.44\%) & 1,321 (23.13\%) & 1,457 (25.51\%) \\
Test & 405 (30.89\%) & 306 (23.34\%) & 300 (22.88\%) & 300 (22.88\%) \\
\hline 
\end{tabularx} \\
\captionof{figure}{\textbf{Classification Class Distribution}}

\includegraphics[width=\linewidth]{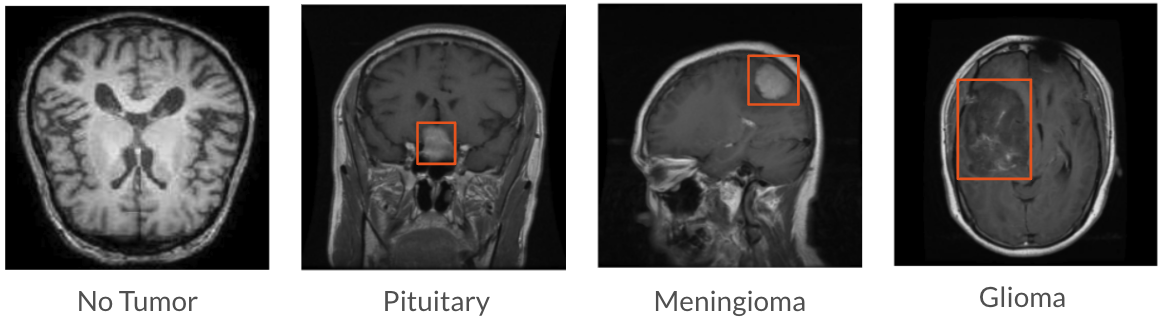}
\captionof{figure}{\textbf{Brain MRI Classification Data}}

\subsection*{Brain MRI Segmentation Dataset}
The LGG Segmentation Dataset contains MRI scans and FLAIR abnormality segmentation masks for 110 patients from The Cancer Genome Atlas (TCGA) lower-grade glioma collection. Images, provided in .tif format with three channels (pre-contrast, FLAIR, post-contrast), and binary masks are organized by patient ID \cite{d2}. The dataset supports studies on tumor shape related to genomic subtypes and patient outcomes. \\

\begin{tabularx}{\linewidth}{|c|X|X|} 
\hline 
Split & No-Tumor & Tumor \\ \hline 
Train & 2,045 (65.07\%) & 1,098 (34.93\%) \\
Test & 511 (65.01\%) & 275 (34.99\%) \\
\hline 
\end{tabularx}\\
\begin{center}
\caption{\bf{Segmentation Class Distribution}}
\end{center}

\subsection*{Brain Tumor Image DataSet: Semantic Segmentation }
The Brain Tumor Image Dataset, designed for the TumorSeg Computer Vision Project, focuses on semantic segmentation with two classes: Tumor (Class 1) and Non-Tumor (Class 0) \cite{d3}. Initially containing 2,146 images, the dataset was adjusted by removing all samples without tumor masks. Notably, the segmentation masks are actually bounding boxes, making this dataset more suitable for object detection tasks in medical image analysis.

\begin{tabularx}{\linewidth}{|X|X|} 
\hline 
Split & Tumor \\ \hline 
Train & 1,501 \\
Test & 215 \\
\hline 
\end{tabularx}\\
\begin{center}
\caption{\bf{Segmentation Box Class Distribution}}
\end{center}

\includegraphics[width=1\linewidth]{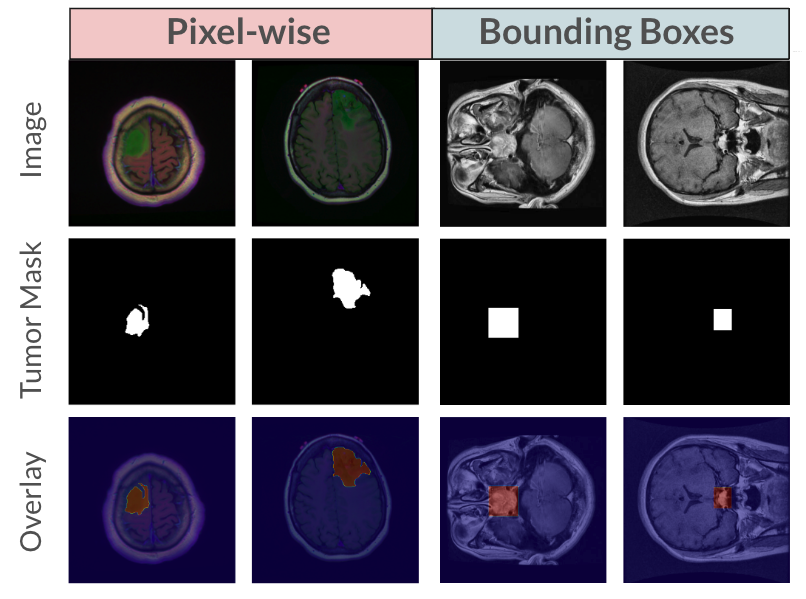}
\caption{\bf{Example of LGG and Box Samples}}

\section*{Image Classification}
\subsection*{Naive Classification}
The first problem we looked at was classification, both binary (tumor or no tumor) and multiclass (what type of tumor). We began investigating detecting the existence of brain tumors in an image by creating a set of baseline, simple models to compare our results to as we increased model complexity throughout the rest of the experiments. For our first baseline model we used logistic regression. Although logistic regression is not typically used on image data, it is simple to implement and trains quickly. The second baseline model we used was a simple Convolutional Nerual Network (CNN). CNNs are more complex than logistic regression and tailored for image-based tasks. CNNs use a kernel (3x3 in our implementation) to capture local context contained in the image. The kernel looks at the area around the current pixel and does a linear combination of the pixels' values, condensing those pixels down to a single feature. The hope is that this output feature encodes information about what is around the target pixel.

\subsubsection*{Methods}
Both of these models were implemented using PyTorch. Logistic Regression was implemented using a single linear layer. Although very simple, this single layer allowed the model to gain simple insights about the images it was seeing. The more complex CNN used two convolutional layers, each with a max pool following it, as well as two fully-connected layers. The only difference between the model for binary and multiclass classification was the output size of the final fully connected layer, changing from 1 output node in the binary case to 4 output nodes for multiclass. Training and evaluation was also handled slightly differently between the binary and multiclass model. All three models (LogReg, Binary CNN, and Multiclass CNN) were trained for 20 epochs. The logistic regression model used a Binary Cross Entropy loss function. The binary CNN used Binary Cross Entryopy with Logits for its loss function. The Multiclass CNN used (non-binary) Cross Entropy Loss. All three models used Stochastic Gradient Descent as their optimizers. As these models were used as simple sanity checks, no programmatic hyperparameter tuning was done. For evaluation, the torchmetrics library was used.

\subsubsection*{Results}

\begin{tabularx}{\linewidth}{|X|X|X|} 
\hline 
Model & Accuracy & AUC \\ \hline 
LogReg & 69.87\% & 0.5151 \\
Binary CNN & 99.54\% & 0.9967 \\
Multiclass CNN & 94.83\% & 0.9663 \\
\hline 
\end{tabularx} \\
\begin{center}
\caption{\bf{Naive Methods Metrics}}
\end{center}

 As was expected, logistic regression did not perform well on this image classification task. The accuracy
is quite low, and the AUC suggests that this classifier is not discernible from a random classifier. This model was only a baseline to which other models
would be compared, so the quality of the classifier is not overly important in our case. The two CNNs improved greatly on the logistic
regression model. After 20 epochs of training, the binary classifier achieved over 99\% accuracy and an AUC very close to 1. The
multiclass CNN, as expected, performed slightly worse on its task than the binary classifier. Given that the two models are very similar
and binary classification is a slightly easier task than multiclass classification, we cannot expect the multiclass CNN to perform as well.
However, the results are still promising. With accuracy close to 95\% and high AUC, we have shown that this model architecture is effective in classifying images of brain tumors. With tweaks to model architecture and more involved hyperparameter tuning, it is likely that these results could be improved. 
The efficiency of these CNNs can be improved as well, which will be shown in the next section*.

\subsection*{ResNet} 
While the CNN performed somewhat satisfactorily, we wanted to explore a more robust and sophisticated architecture to capture the complex patterns in high-dimensional image data. ResNet or Residual Network is known for its deep network capabilities and effectiveness in handling vanishing gradients\cite{he2015deep}. Its success covers various image classification tasks, especially in scenarios involving complex visual data like medical images, making it an ideal candidate for this project.

\subsubsection*{Methods}
Our implementation of ResNet tailored specifically for brain tumor image classification involved constructing a custom ResNet architecture. This was essential for handling the unique challenges presented by medical imaging data. Central to the ResNet architecture are the residual blocks, which facilitate the training of deeper networks without suffering from vanishing gradients. Each residual block includes skip connections that allow inputs to bypass one or more layers. These connections help mitigate the vanishing gradient problem by facilitating direct gradient flow during backpropagation, which is crucial for maintaining performance integrity across many layers.

\subsubsection*{Network Architecture}
Our custom ResNet model starts with an initial convolutional layer and max pooling to prepare the input for a series of residual blocks, each doubling the number of filters and reducing the spatial dimensions through striding. The architecture culminates in average pooling and a fully connected layer tailored to the specific classification task—binary or multi-class—allowing for flexible use across different types of data. Additionally, the model is dynamically constructed to adjust the network's depth and capacity as needed, using a method that configures each layer for optimal downsampling or dimension maintenance. We initialized the model to suit the MRI image characteristics, employing Adam as the optimizer for its advantages in managing sparse gradients and adaptive learning rates. This configuration is particularly advantageous for handling the complex and varied nature of medical imaging datasets.

\subsubsection*{Hyperparameter Tuning}
Hyperparameter tuning was a crucial aspect of our model development. We experimented with different optimizers and found Adam to be superior in our context due to its faster convergence properties. The initial use of a learning rate scheduler was intended to adjust the learning rate based on training progress, but it was removed after observing a decrease in performance, indicating potential overfitting or inadequate learning rate adjustments. The final learning rate of 0.001 was selected after several trials to ensure a balance between training speed and convergence stability.

\subsubsection*{Results}
Implementing ResNet improved some of our classification results: \\

\begin{tabularx}{\linewidth}{|X|X|X|} 
\hline 
ResNet Variant & Accuracy & AUC \\ \hline 
Binary & 97.71\% & 0.968 \\
Multiclass & 94.35\% & 0.996 \\
\hline 
\end{tabularx} \\
\caption{\bf{ResNet Metrics}}

While the performance of ResNet shows only a slight improvement in accuracy and AUC when compared to the naive CNN, it was trained in half the number of epochs. This illustrates ResNet's efficiency and robustness in processing complex image data, especially in regards to the AUC for multiclass. This performance boost is attributed to the deeper network architecture's ability to learn detailed and hierarchical features from the brain MRI images, which are crucial for accurate classification. The use of ResNet in our project not only improved classification accuracy but also highlighted the potential of deep learning in enhancing diagnostic processes through advanced image recognition capabilities.

\section*{Semantic Segmentation}

As we transition from holistic image classification, where the entire image is assigned a single label, to pixel-wise semantic segmentation, our approach to image analysis becomes more granular and complex. In holistic classification, the primary goal is determining whether an image belongs to one specific class. While this may be initially useful, we may want more localized information about the image. 
Semantic segmentation addresses this limitation by assigning a class label to each pixel, effectively enabling a detailed map of various tissues and anomalies within a single image. This shift changes our targets—from a class label per image to a class label per pixel—and introduces new challenges in computational complexity and accuracy.

 In pixel-wise classification, traditional metrics like overall accuracy become less informative due to the high prevalence of background or non-tumor pixels overwhelming the minority class of interest (tumor pixels). Therefore, we adopt Intersection over Union (IoU)\cite{wang2023revisiting} as a more suitable metric for this task. IoU provides a balanced measure by evaluating the overlap per class between the predicted and actual segments relative to their combined area, thus effectively handling class imbalances and focusing on the precision of boundary delineation.
\subsection*{Related Work}
In the initial stages of our project, we planned to explore various model architectures for image segmentation. However, given the complexities and the intensive knowledge required for each, coupled with our project's limited timeframe, we opted to focus solely on a single, impactful architecture.

\begin{figure*}[t]  
\centering
\includegraphics[scale=0.35]{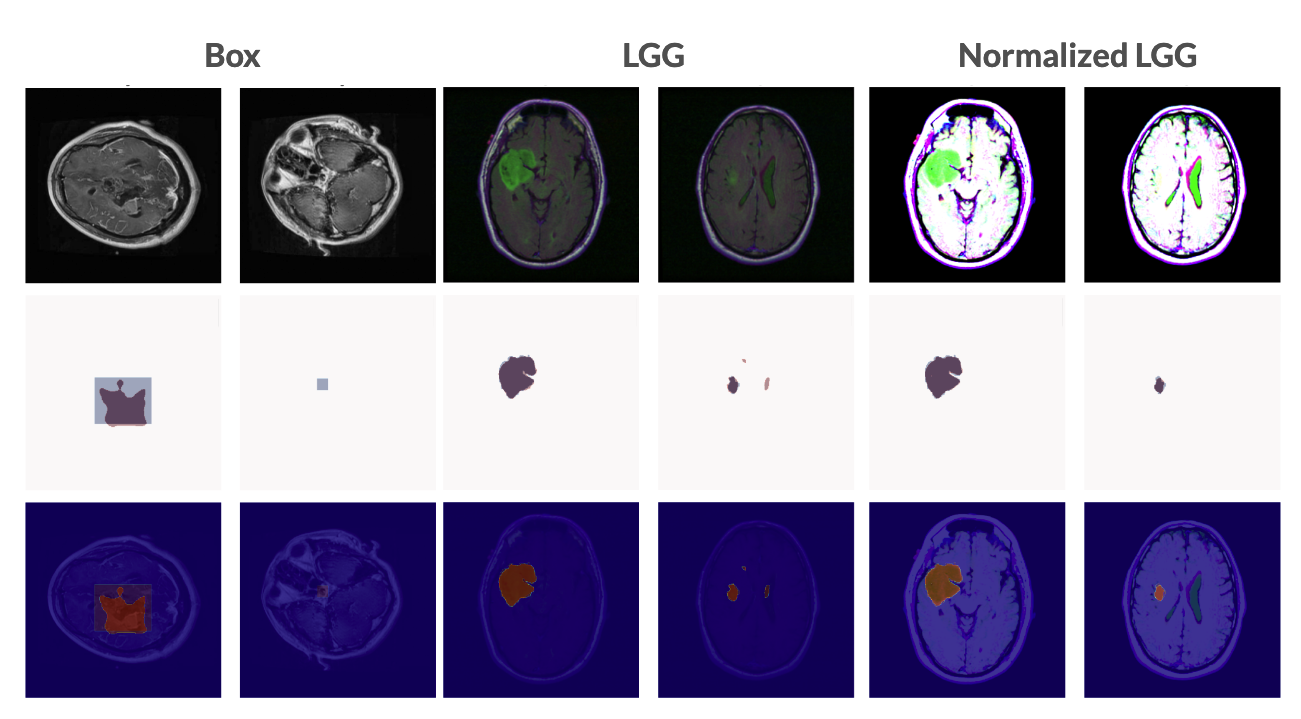}
\caption{\textbf{Results of U-Net training on the test splits of various datasets. Center image shows overlap between label (blue) and prediction (red).}}
\end{figure*}

 We selected the U-Net\cite{ronneberger2015unet} architecture for its innovative approach. Traditional segmentation methods like the sliding window technique require processing individual patches around each pixel independently, a method that is extremely computationally inefficient and often overlooks important contextual relationships within the image. U-Net revolutionized this by adopting an encoder-decoder architecture, commonly used in language processing, to analyze the entire image holistically. This significantly reduced the need for redundant processing of overlapping patches, enhancing efficiency and effectiveness in capturing detailed image contexts.

\textbf{Encoder-Decoder Structure:} In the U-Net architecture, the encoder uses a combination of convolutional layers and max-pooling operations to reduce the spatial dimensions of an input image while increasing the depth of feature maps. This encoding process effectively narrows the representation of features into a deeper, more abstract form, capturing essential contextual information at multiple scales. Conversely, the decoder works to reverse this process by progressively expanding the compressed feature maps. It attempts to reconstruct the original spatial dimensions from this high-level latent representation, translating it back into detailed segmentation masks that match our target structures. This decoding is vital for restoring high-resolution details necessary for accurate medical diagnostics. 

\textbf{Skip Connections:}\cite{wu2020skip} An essential feature of U-Net, skip connections are critical in enhancing the decoding process. As the encoder compresses the image, it extracts and condenses high-level features, sometimes resulting in the loss of finer details. Skip connections address this by directly linking the detailed, information-rich feature maps from the encoder's early layers to each step of the decoding process. This mix of detailed early-stage data with high-level abstracted features allows the decoder to reconstruct the segmentation map more accurately. Such integration is especially beneficial in medical imaging, where preserving precise anatomical details is crucial for accurate diagnostics.
\subsection*{Modifications}

\textbf{Simplification for Smaller Images:} Unlike the original U-Net, which used a tiling approach to manage large, high-resolution images, we adapted the architecture to suit smaller, uniformly sized images (320x320x3). This decision avoids the complexities associated with tiling and cropping, which in the original U-Net included in both training objectives and preprocessing steps to ensure accuracy around the edges of tiles. By using smaller images, we could streamline the preprocessing and reduce the computational overhead, allowing the focus to remain on accurate segmentation without the need for edge weighting or extensive padding.

\textbf{Binary Classification:} To further simplify the model, we shifted the output goal to a binary classification system, where each pixel is classified simply as 'Tumor' or 'No Tumor'. This change reduced the complexity of the output layer, focusing on the critical task at hand and making the training process more straightforward. We used binary cross entropy as the loss function to support this two-class system directly.

\textbf{Optimization and Hyperparameter Tuning:} We employed Stochastic Gradient Descent (SGD) to optimize the model, inspired by its usage in the original U-Net paper. Additionally, recognizing the constraints posed by our limited computational resources, we utilized a combination of Bayesian Optimization\cite{frazier2018tutorial} and K-fold cross-validation\cite{yates2022cross} to tune the model's hyperparameters efficiently. Bayesian Optimization provided a more adaptive approach than traditional grid search for finding optimal parameters, allowing us to iteratively test and refine parameters like batch size and learning rate scheduler settings. K-fold cross-validation helped assess the model's performance more robustly across different data splits, thus reducing biases that could affect our evaluation of the model's effectiveness.

\subsection*{Results}

Our implementation of U-Net performed quite well! After training our model for 50 epochs and implementing a linear decay of our initial learning rate by a factor of $0.1$ every 20 epochs, we achieved the following results across our datasets.

\vspace{0.2cm}

\begin{tabularx}{\linewidth}{|c|X|X|} 
\hline 
Dataset & IoU & AUC \\ \hline 
Box & 57.6\% & 0.975 \\
LGG & 70.3\% & 0.996 \\
Normalized LGG & 70.9\% & 0.996 \\
\hline 
\end{tabularx}
\begin{center}
\caption{\bf{Segmentation Results}}
\end{center}

Our U-Net implementation demonstrated promising results. Despite the complexity of semantic segmentation, an IoU of 70\% is commendable, complemented by a high AUROC, indicating effective minimization of false positives. Performance was strong on the LGG dataset but less so within the BOX dataset, though false positives remained low.

Future work will focus on expanded hyperparameter tuning and exploring data augmentation techniques like cropping and shifting to enhance model robustness and potentially implement the tiling approach described in the original U-Net paper.

\section*{Anchor-based Object Detection}
\subsection*{Related Work}

Object detection is a popular computer vision technique used to identify specific elements within images. This method involves drawing a bounding box around each object, defined by four corner coordinates, and assigning a label to the box that indicates the class of the object it encloses. Unlike image segmentation, object detection must handle the detection of multiple objects of one or more classes at various scales and locations within a single image. To accommodate these challenges, the formulation of the problem is carefully adjusted to ensure precise and efficient detection.

More specifically, anchor-based object detection starts by defining a variety of anchor boxes, each varying in size and aspect ratio, across an entire image \cite{ren2015fasterrcnnarxiv}. These anchors are designed to potentially encompass all regions and scales of objects present in the image. During preprocessing, each image is labeled by examining these predefined anchor boxes to determine their overlap with the ground truth bounding boxes. Each anchor is then assigned a label: either 'no object' or one of the specific object classes \cite{ren2015fasterrcnnarxiv}.

The model's initial task is to classify each anchor box for every sample image. However, classification alone is insufficient for completing the object detection process, as it is rare for an anchor box to perfectly align with a ground truth bounding box. Therefore, for anchors classified as containing an object, the model must also predict the necessary adjustments to better fit the ground truth. This involves calculating the required shifts in the anchor's center coordinates, as well as adjustments to its width and height. These offset values are predicted through regression, adding another layer of complexity to the model's tasks. This overall approach was adapted for the detection of tumors within brain MRI images.

\subsection*{Methods}

Images and their ground truth bounding boxes were resized to 3x256x256. Anchor boxes were configured with image scales of 0.1, 0.175, and 0.3, aspect ratios of 2:1, 1:1, and 1:2 (reflecting the approximately circular nature of most tumors), and a feature map size of 32x32. This configuration resulted in a total of 9216 anchor boxes. With these parameters, 99.5\% of the training bounding boxes achieved an Intersection* over Union (IoU) greater than 0.3 with at least one anchor. For IoUs of 0.5 and 0.75, the coverage rates were 96.7\% and 45.1\%, respectively, indicating that these anchors provided adequate coverage of the ground truths within the training set.

The labeling of each anchor for each sample adhered to the following criteria: any box with an IoU over 0.5 was considered positive, while all others were labeled as negative. In cases where a sample lacked a positive anchor, the anchor with the highest overlap (provided that the IoU was over 0.3) was assigned a positive label. This policy ensured that 99.5\% of the images had at least one positive anchor. For all positive anchors, offsets for the center coordinates and adjustments in width and height were calculated as regression targets. 

\includegraphics[width=1\linewidth]{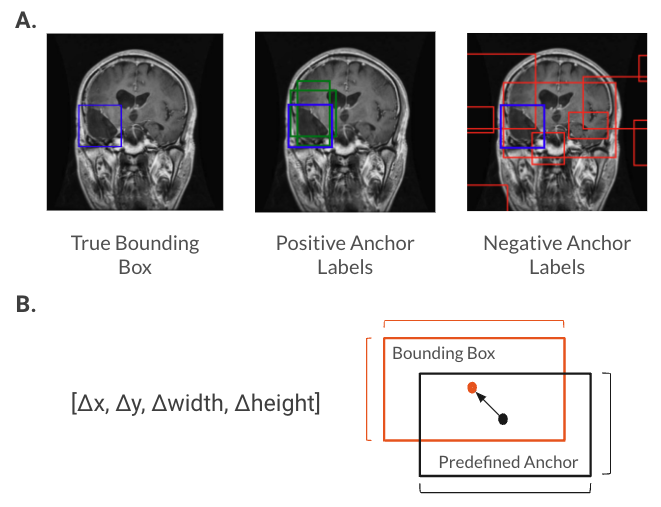}
\caption{\bf{(A) Anchor classification (B) Anchor regression}}

The selected model, EfficientDet, employs an EfficientNet backbone for robust feature encoding, combined with a bi-directional feature pyramid network (BiFPN) for multi-scale feature fusion \cite{tan2020efficientdet}. This architecture also includes two key predictors: a classifier to determine anchor labels and a regressor for predicting anchor offsets. We utilized EfficientNet-B0, the smallest variant in the EfficientNet series with 3M parameters, to encode images into multi-resolution features \cite{tan2019efficientnet}. 64 channels of encoded features following dimension reducing convolutions (64x32x32, 64x16x16, 64x8x8, 64x4x4, and 64x2x2) are fed into the BiFPN, which consists of 3 BiFPN layers. In each BiFPN layer, features from different resolutions are fused in a weighted manner, allowing for interaction between lower and higher resolution data. The features from the 64x32x32 resolution are input into both the classifier and regressor, each consisting of three convolutional layers with ReLU activations, maintaining the same dimensions. The classifier and regressor output a single prediction (logit of true class) and four offset values per anchor, respectively. The resulting model featured 4M trainable parameters.

In our object detection experiments, we tested a range of hyperparameters to optimize performance. Over the course of 50 epochs, we experimented with two initialization strategies for EfficientNet-B0: using pretrained weights and training from scratch. We evaluated three different learning rates: 0.1, 0.01, and 0.001. For the regression tasks, we compared the mean squared error (MSE) loss function with SmoothL1Loss, while for classification, we tested binary cross-entropy (BCE), weighted BCE (see formula below), and focal loss (alpha=0.1, gamma=2.0) \cite{lin2017focalarxiv}. The latter two were considered as a means of addressing class imbalance. Additionally, two optimizers were used in our trials: Adam and Stochastic Gradient Descent (SGD). The batch size was fixed at 16 samples. These variations were chosen to determine the most effective combination for our object detection model.

\[
\text{weighted\_BCE} = -w_{\text{pos}} \cdot y \cdot \log(\hat{y}) - w_{\text{neg}} \cdot (1 - y) \cdot \log(1 - \hat{y})
\]

 where: \( w_{\text{pos}} \) is calculated as the ratio of negative samples to total samples, and \( w_{\text{neg}} \) is \( 1 - w_{\text{pos}} \). \\

In our study, unlike typical scenarios involving multiple objects of the same or different classes within a single image, our dataset uniquely featured just one ground truth bounding box per image, labeled as "tumor." This simplicity significantly streamlined our performance evaluation process. For each image, we focused solely on the anchor with the highest predicted probability. The coordinates of this anchor were adjusted using the regressor's outputs, and the IoU was then calculated between the adjusted prediction and the actual ground truth bounding box. Additionally, to evaluate the effectiveness of our anchor classification, we computed the AUC for each sample.

\subsection*{Results and Discussion}
Overall, this approach to the problem proved less effective compared to our semantic segmentation methods. We froze the regressor and focused primarily on anchor classification, as identifying which anchors contain tumors is essential before adjusting them to fit the bounding boxes. Throughout most experimental trials, gradients tended to explode by the final epoch, complicating the assessment of optimal hyperparameters. However, several key observations emerged that could inform future research.

The Adam optimizer frequently led to gradient explosions, particularly within the first 10 epochs across all tested learning rates and loss functions. In contrast, the SGD optimizer facilitated more stable training dynamics. The weighted BCE loss function also demonstrated greater stability compared to other loss options. Utilizing a pretrained backbone did not yield a notable advantage. The highest validation IoU and AUC recorded were 0.075 and 0.801, respectively, achieved with a pretrained backbone, SGD optimizer at a learning rate of 0.01, and weighted BCE loss—still a significant underperformance compared to semantic segmentation.

The model and training approach have a vast hyperparameter space that warrants further exploration. Future experiments should consider scaling up the model size, including the backbone, BiFPN, and prediction heads, following the guidelines proposed by 2020 M. Tan et al. The limited number of samples is another significant issue since object detection models typically require extensive datasets for effective training. Addressing class imbalance in predictions should also be a priority in subsequent studies.

\section*{Conclusion}
This study has successfully reinforced the significant potential of deep learning techniques in the automated detection and segmentation of brain tumors from MRI scans. By utilizing a variety of models, CNNs, and ResNet, alongside advanced methods like U-Net for segmentation, we have shown marked improvements in both accuracy and efficiency over traditional methods. The application of object detection models requires further exploration due to the encountered challenges described in the results. Future work will aim to further refine these models, explore more complex architectures, and expand the datasets for training to improve the generalizability and accuracy of the systems. This project lays a foundation for the integration of deep learning into clinical practice with aims of improving patient outcomes through more precise and timely diagnosis.

\section*{Contributions}
Jake Lynn - Naive Classification and Base CNN; \\
Jack Krolik - Binary and Multi Classification ResNet; \\
John Henry Rudden - Dataset, Training, and Visualization Pipelines, Semantic Segmentation; \\
Dmytro Vremenko - Anchor-Based Object Detection 

\end{multicols}
\begin{multicols}{2}
    \bibliographystyle{unsrt}
    \bibliography{references}

\end{multicols}

\end{document}